\documentclass[10pt,twocolumn,letterpaper]{article}
\usepackage{cvpr}              

\usepackage{dsfont}
\usepackage{graphicx}
\usepackage{amsmath}
\usepackage{amssymb}
\usepackage{booktabs}
\usepackage{enumitem} 
\usepackage[title]{appendix}

\usepackage[pagebackref,breaklinks,colorlinks]{hyperref}

\usepackage[capitalize]{cleveref}
\crefname{section}{Sec.}{Secs.}
\Crefname{section}{Section}{Sections}
\Crefname{table}{Table}{Tables}
\crefname{table}{Tab.}{Tabs.}

\def\cvprPaperID{2978} 
\def\confName{CVPR}
\def\confYear{2023}

\begin{document}

\title{Paint by Example: Exemplar-based Image Editing with Diffusion Models}
\author{
{Binxin Yang{$^{1}$}}\thanks{Author did this work during his internship at Microsoft Research Asia.} \qquad Shuyang Gu{$^{2}$} \qquad  Bo Zhang{$^{2}$} \qquad  Ting Zhang{$^{2}$} \qquad Xuejin Chen{$^{1}$} \\
\qquad Xiaoyan Sun{$^{1}$} \qquad Dong Chen{$^{2}$} \qquad Fang Wen{$^{2}$} \qquad \\
\normalsize
$^{1}$\    University of Science and Technology of China ~~ $^{2}$\,Microsoft Research Asia\\
}

\twocolumn[{%
\renewcommand\twocolumn[1][]{#1}%
\maketitle
\begin{center}
    \centering
    \vspace{-0.5cm}
    \captionsetup{type=figure}
    \includegraphics[width=1\textwidth]{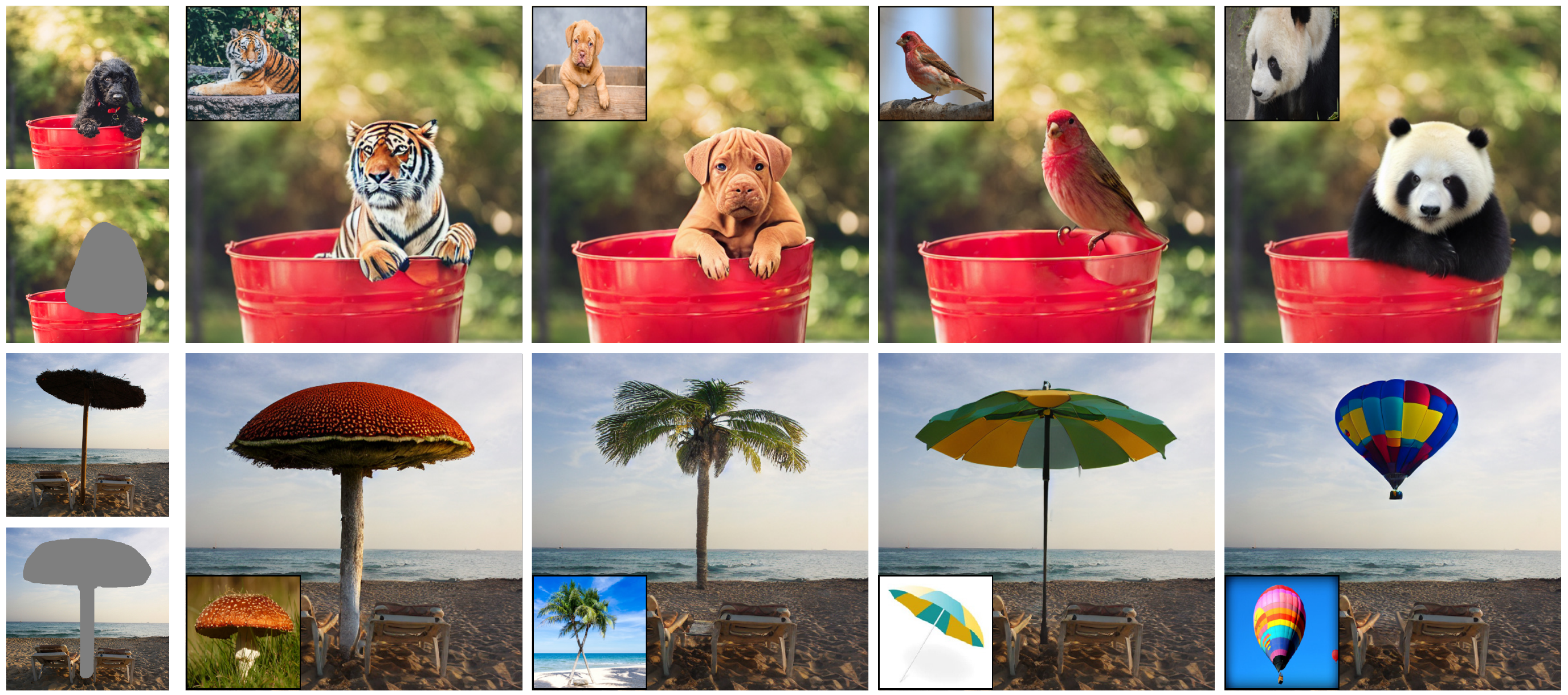}
    \captionof{figure}{Paint by example. Users are able to edit a scene by painting with a conditional image. Our approach can automatically alter the reference image and merge it into the source image, and achieve a high-quality result.}
    \label{fig:teaser}
\end{center}%
}]

{
  \renewcommand{\thefootnote}%
    {\fnsymbol{footnote}}
  \footnotetext[1]{Author did this work during his internship at Microsoft Research Asia.}
}

\begin{abstract}
Language-guided image editing has achieved great success recently. In this paper, for the first time, we investigate exemplar-guided image editing for more precise control. 
We achieve this goal by leveraging self-supervised training to disentangle and re-organize the source image and the exemplar. However, the naive approach will cause obvious fusing artifacts. 
We carefully analyze it and propose an information bottleneck and strong augmentations to avoid the trivial solution of directly copying and pasting the exemplar image. 
Meanwhile, to ensure the controllability of the editing process, we design an arbitrary shape mask for the exemplar image and leverage the classifier-free guidance to increase the similarity to the exemplar image. 
The whole framework involves a single forward of the diffusion model without any iterative optimization.
We demonstrate that our method achieves an impressive performance and enables controllable editing on in-the-wild images with high fidelity. The code and pretrained models are available at \url{https://github.com/Fantasy-Studio/Paint-by-Example}.
\end{abstract}

\section{Introduction}
\label{sec:intro}

Creative editing for photos has become a ubiquitous need due to the advances in a plethora of social media platforms. AI-based techniques~\cite{liu2021generative} significantly lower the barrier of fancy image editing that traditionally requires specialized software and labor-intensive manual operations. Deep neural networks can now produce compelling results for various low-level image editing tasks, such as image inpainting~\cite{suvorov2022resolution,yu2018generative}, composition~\cite{zhang2021deep,xue2022dccf,niu2021making}, colorization~\cite{zhang2016colorful,zhang2019deep,saharia2022palette} and aesthetic enhancement~\cite{deng2018aesthetic,chen2018deep}, by learning from richly available paired data. A more challenging scenario, on the other hand, is semantic image editing, which intends to manipulate the high-level semantics of image content while preserving image realism. Tremendous efforts~\cite{ling2021editgan,shen2020interpreting,alaluf2022hyperstyle,roich2022pivotal,shen2021closed,bau2020semantic} have been made along this way, mostly relying on the semantic latent space of generative models, \eg, GANs~\cite{goodfellow2020generative,karras2019style,zhang2022styleswin}, yet the majority of existing works are limited to specific image genres.

Recent large-scale language-image (LLI) models, based on either auto-regressive models~\cite{yu2022scaling,gafni2022make} or diffusion models~\cite{ramesh2022hierarchical,saharia2022photorealistic,rombach2021highresolution,gu2022vector}, have shown unprecedented generative power in modeling complex images. These models enable various image manipulation tasks~\cite{hertz2022prompt,ruiz2022dreambooth,kawar2022imagic,avrahami2022blended} previously unassailable, allowing image editing for general images with the guidance of text prompt. 
However, even the detailed textual description inevitably introduces ambiguity and may not accurately reflect the user-desired effects; indeed, many fine-grained object appearances can hardly be specified by the plain language. Hence, it is crucial to develop a more intuitive approach to ease fine-grained image editing for novices and non-native speakers.

In this work, we propose an \emph{exemplar-based image editing} approach that allows accurate semantic manipulation on the image content according to an exemplar image
provided by users or retrieved from the database. As the saying goes, ``a picture is worth a thousand words''. We believe images better convey the user's desired image customization in a more granular manner than words.  This task is completely different from image harmonization~\cite{tsai2017deep,guo2021image} that mainly focuses on color and lighting correction when compositing the foreground objects, whereas we aim for a much more complex job: semantically transforming the exemplar, \eg, producing a varied pose, deformation or viewpoint, such that the edited content can be seamlessly implanted according to the image context. 
In fact, ours automates the traditional image editing workflow where artists perform tedious transformations upon image assets for coherent image blending.

To achieve our goal, we train a diffusion model~\cite{ho2020denoising,song2020score} conditioned on the exemplar image. Different from text-guided models, the core challenge is that it is infeasible to collect enough triplet training pairs comprising source image, exemplar and corresponding editing ground truth. One workaround is to randomly crop the objects from the input image, which serves as the reference when training the inpainting model. The model trained from such a \emph{self-reference} setting, however, cannot generalize to real exemplars, since the model simply learns to copy and paste the reference object into the final output. We identify several key factors that circumvent this issue. The first is to utilize a generative prior. Specifically, a pretrained text-to-image model has the ability to generate high-quality desired results, we leverage it as initialization to avoid falling into the copy-and-paste trivial solution.
However, a long time of finetuning may still cause the model to deviate from the prior knowledge and ultimately degenerate again. Hence, we introduce the information bottleneck for self-reference conditioning in which we drop the spatial tokens and only regard the global image embedding as the condition. In this way, we enforce the network to understand the high-level semantics of the exemplar image and the context from the source image, thus preventing trivial results during the self-supervised training.
Moreover, we apply aggressive augmentation on the self-reference image which can effectively reduce the training-test gap.  

We further improve the editability of our approach in two aspects. One is that our training uses irregular random masks so as to mimic the casual user brush used in practical editing. We also prove that classifier-free guidance~\cite{ho2022classifier} is beneficial to boost both the image quality and the style resemblance to the reference. 

\begin{figure*}[h]
\centering
\vspace{-0.8cm}
\includegraphics[width=1\textwidth]{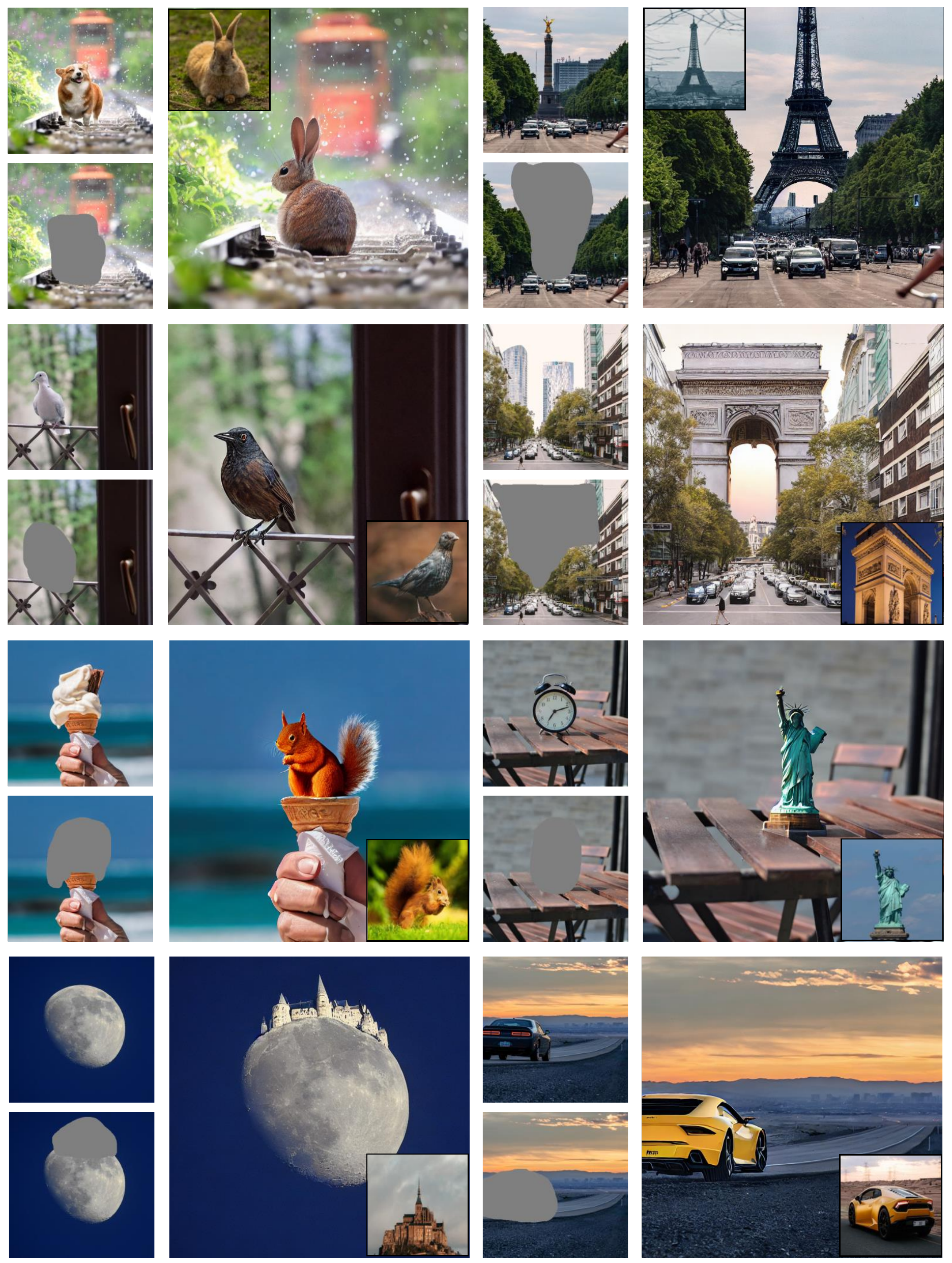}
\caption{More visual results. Our approach can handle a wide variety of reference images and source images.  }
\vspace{-0.8cm}
\label{fig:supp_3}
\end{figure*}

To the best of our knowledge, we are the first to address this \emph{semantic image composition} problem where the reference is semantically transformed and harmonized before blending into another image, as shown in Figure~\ref{fig:teaser} and Figure~\ref{fig:supp_3}. Our method shows a significant quality advantage over prior works in a similar setting. Notably, our editing just involves a single forward of the diffusion model without any image-specific optimization, which is a necessity for many real applications. To summarize, our contributions are as follows:
\begin{itemize}[leftmargin=*]
    \itemsep=-0.9mm
    \item We propose a new image editing scenario, which semantically alters the image content based on an exemplar image. This approach offers fine-grained control while being convenient to use.  
    \item We solve the problem with an image-conditioned diffusion model trained in a self-supervised manner. We propose a group of techniques to tackle the degenerate challenge.
    \item Our approach performs favorably over prior arts for in-the-wild image editing, as measured by both quantitative metrics and subjective evaluation.
\end{itemize}

\section{Related Work}
\label{sec:related_work}

\noindent
\textbf{Image composition.}
Cutting the foreground from one image and pasting it on another image into a realistic composite is a common and widely used operation in photo editing.
Many methods~\cite{cohen2006color, jia2006drag, tao2010error,sunkavalli2010multi,tsai2017deep,cong2020dovenet,cun2020improving,reinhard2001color,perez2003poisson} have been proposed focusing on image harmonization to make the composite look more realistic.
Traditional methods~\cite{cohen2006color, jia2006drag, tao2010error} tend to extract handcrafted features to match the color distribution. Recent works~\cite{chen2019toward,guo2021intrinsic} leverage deep semantic features to improve the robustness. 
A more recent work DCCF~\cite{xue2022dccf} proposes four human comprehensible neural filters in a pyramid manner and achieves a state-of-the-art color harmonization result. However, they all assume that the foreground and the background are semantically harmonious and only adjust the composite in the low-level color space while keeping the structure unchanged.
In this paper, we target at semantic image composition, taking the challenging semantic inharmony into consideration.

\noindent
\textbf{Semantic image editing.}
Semantic image editing, to edit the high-level semantics of an image, is of great interest in the vision and graphics community due to its potential applications. A steadily-developed line of work~\cite{shen2020interpreting, shen2021closed, richardson2021encoding, alaluf2022hyperstyle,bau2020semantic} carefully dissects the GAN's latent space, aiming to find semantic disentangled latent factors. 
Whereas other research efforts leverage the discriminant model like attribute classifier~\cite{gao2021high, hou2022guidedstyle} or face recognition~\cite{li2019faceshifter, shen2018faceid} model to help disentangle and manipulate images.
Another popular direction of
works~\cite{gu2019mask, bau2020semantic, ling2021editgan,zhang2020cross,zhou2021cocosnet,wang2022pretraining} utilize semantic mask to control the editing. 
Yet most existing methods are limited to specific image genres, such as face, car, bird, cat \etc. 
In this work, we focus on introducing a model that works for general and complex images in a high-precision manner.

\noindent
\textbf{Text-driven image editing.}
Among the various kinds of semantic image editing, text-guided image editing has been attracting increasing attention recently. Early works~\cite{patashnik2021styleclip,abdal2022clip2stylegan,xia2021tedigan,gal2022stylegan,bau2021paint}
leverage pretrained GAN generators~\cite{karras2020analyzing} and text encoders~\cite{radford2021learning} to progressively optimize the image according to the text prompt.
However, these GAN-based manipulation approaches struggle on editing images of complex scenes or various objects due to the limited modeling capability of GANs. 
The rapid rise and development of diffusion models~\cite{ramesh2021zero, ramesh2022hierarchical, saharia2022photorealistic} have shown powerful capability in synthesizing high-quality and diverse images. Many works~\cite{liu2021more,kim2022diffusionclip,ruiz2022dreambooth,kawar2022imagic,nichol2021glide, avrahami2022blended,hertz2022prompt} exploit diffusion models for text-driven image editing.
For example,
DiffusionCLIP~\cite{kim2022diffusionclip}, dreambooth~\cite{ruiz2022dreambooth}, and Imagic~\cite{kawar2022imagic}  finetune the diffusion models case-specifically for different text prompts.
Blended Diffusion~\cite{avrahami2022blended} proposes a multi-step blended process to perform local manipulation using a user-provided mask.
While these methods achieve remarkably impressive results, we argue that the language guidance still lacks precise control, 
whereas images can better express one's concrete ideas.
As such in this work we are interested in exemplar-based image editing.

\section{Method}
\label{sec:method}
We target at exemplar-based image editing that automatically merges the reference image (either retrieved from a database or provided by users) into a source image in a way that the merged image looks plausible and photo-realistic.
Despite the recent remarkable success of text-based image editing, it is still difficult to use mere verbal descriptions to express complex and multiple ideas.
While images, on the other hand, could be a better alternative for conveying people's intentions, as the proverb says: ``a picture is worth a thousand words".
 
Formally, denote the source image as $\mathbf{x}_s\in \mathbb{R}^{H\times W\times 3}$, with $H$ and $W$ being the width and height respectively.
The edit region could be a rectangular or an irregular shape (at least connected) and is represented as 
a binary mask $\mathbf{m}\in \{0, 1\}^{H\times W}$ where value $1$ specifies the editable positions in $\mathbf{x}_s$.
Given a reference image $\mathbf{x}_r \in \mathbb{R}^{H'\times W'\times 3}$ containing the desired object,
our goal is to synthesize an image $\mathbf{y}$ from $\{\mathbf{x}_s,\mathbf{x}_r, \mathbf{m}\}$, so that the region where $\mathbf{m}=0$ remains as same as possible to the source image $\mathbf{x}_s$, while the region where $\mathbf{m}=1$ depicts the object as similar to the reference image $\mathbf{x}_r$ and fits harmoniously.

This task is very challenging and complex because it implicitly involves several non-trivial procedures.
Firstly, the model requires understanding the object in the reference image, capturing both its shape and texture while ignoring the noise from the background.
Secondly, it is critical to enable synthesizing a transformed view of the object (different pose, different size, different illumination \etc) that fits in the source image nicely.
Thirdly, the model needs to inpaint the area around the object to generate a realistic photo, showing a smooth transition across the merging boundary.
Last, the resolution of the reference image may be lower than the edit region. The model should involve super-resolution in the process.



\subsection{Preliminaries}
\label{method:naive}

\noindent
\textbf{Self-supervised training.}
It is impossible to collect and annotate paired data, \ie $\{(\mathbf{x}_s,\mathbf{x}_r, \mathbf{m}), \mathbf{y}\}$, for the training of exemplar-based image editing. It may take great expense and huge labor to manually paint reasonable output.
Thus, we propose to perform self-supervised training. Specifically, given an image and the bounding box of an object in the image, to simulate the training data, we use the bounding boxes of the object as the binary mask $\mathbf{m}$. We directly regard the image patch in the bounding box of the source image as the reference image $\mathbf{x}_r = \mathbf{m}\odot \mathbf{x}_s$. Naturally, the image editing result should be the original source image $\mathbf{x}_s$.
As such, our training data is composed of $\{(\bar{\mathbf{m}}\odot \mathbf{x}_s, \mathbf{x}_r, \mathbf{m}), \mathbf{x}_s\}$,  where $\bar{\mathbf{m}} = \mathds{1} - \mathbf{m}$ stands for the complementary of the mask $\mathbf{m}$, and $\mathds{1}$ represents the all-ones matrix.

\noindent
\textbf{A naive solution.}
Diffusion models have achieved notable progress in synthesizing unprecedented image quality and have been successfully applied to 
many text-based image editing works~\cite{GLIDE,kim2022diffusionclip,ruiz2022dreambooth,kawar2022imagic}.
For our exemplar-based image editing task, a naive solution is to directly replace the text condition with the reference image condition.

Specifically, the diffusion model generates image $\mathbf{y}$ by gradually reversing a Markov forward process. Starting from $\mathbf{y}_0 = \mathbf{x}_s$, the forward process yields a sequence of increasing noisy images $\{\mathbf{y}_t | t \in [1, T]\}$, where $\mathbf{y}_t = \alpha_t \mathbf{y}_0 + (1-\alpha_t)\mathbf{\epsilon}$, $\epsilon$ is the Gaussian noise, and $\alpha_t$ decreases with the timestep $t$. For the generative process, the diffusion model progressively denoises a noisy image from the last step given the condition $\mathbf{c}$ by minimizing the following loss function:
\begin{equation}
    \mathcal{L} = \mathbb{E}_{t, \mathbf{y}_0,\mathbf{\epsilon}}\left\|\mathbf{\epsilon}_\theta(\mathbf{y}_t, \bar{\mathbf{m}}\odot \mathbf{x}_s, \mathbf{c}, t) - \mathbf{\epsilon}\right\|^2_2.
\end{equation}
For text-guided inpainting models, the condition $\mathbf{c}$ is the given text and is usually processed by a pretrained CLIP~\cite{radford2021learning} text encoder, outputting $77$ tokens.
Likewise, a naive solution is to directly replace it with CLIP image embeddings. We leverage the pretrained CLIP image encoder outputting $257$ tokens, including 1 class tokens and 256 patch tokens, denoted as $\mathbf{c} = \text{CLIP}_{\text{all}}(\mathbf{x}_r)$.

\begin{figure}[t]
\centering
\vspace{-0.1cm}
\includegraphics[width=1.0\columnwidth]{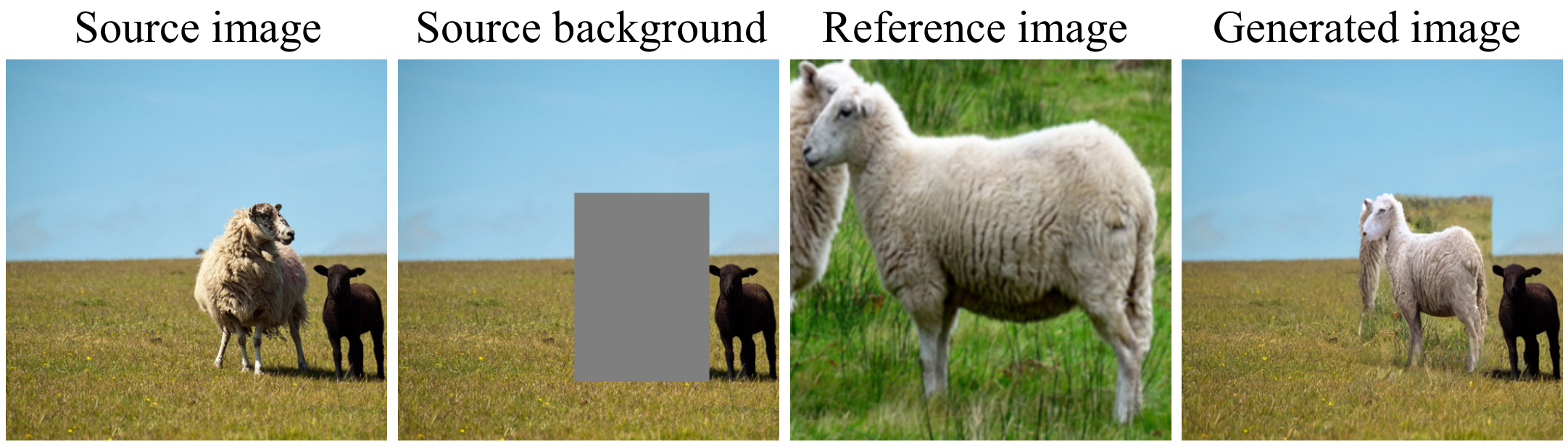}
\vspace{-0.6cm}
\caption{Illustration of the copy-and-paste artifacts of the naive solution. The generated image is extremely unnatural.}
\vspace{-0.3cm}
\label{fig:issue}
\end{figure}

This naive solution converges well on the training set. However, when we apply to test images, we found that the generated result is far from satisfactory. 
There exist obvious copy-and-paste artifacts in the edit region, making the generated image extremely unnatural, as illustrated in Figure~\ref{fig:issue}.
We argue that this is because, under the naive training scheme,  the model learns a trivial mapping function: $ \bar{\mathbf{m}} \odot \mathbf{x}_s + \mathbf{x}_r = \mathbf{x}_s$. 
It impedes the network to understand the content in the reference image and the connection to the source image, leading to failure generalization to test scenarios where the reference image is given arbitrarily but not the patch from the original image.

\noindent
\textbf{Our motivation.}
How to prevent the model from learning such a trivial mapping function and facilitate model understanding in a self-supervised training manner is a challenging problem.
In this paper, we propose three principles. 
1) We introduce the information bottleneck to force the network to understand and regenerate the content of the reference image instead of just copy.
2) We adopt strong augmentation to mitigate the train-test mismatch issue. This helps the network not only learn the transformation from the exemplar object, but also from the background.
3) Another critical feature for exemplar-based image editing is controllability. We enable control over the shape of the edit region and the similarity degree between the edit region and the reference image.

\subsection{Model Designs}
\label{method:ours}
\subsubsection{Information Bottleneck}

\noindent
\textbf{Compressed representation.}
We re-analyze the difference between text condition and image condition. 
For text condition, the model is naturally compelled to learn semantics as text is an intrinsically semantic signal.
In regards to image condition, it is very easy to remember instead of understanding the context information and copying the content, arriving at the trivial solution.
To avoid this, we intend to increase the difficulty of reconstructing the mask region by compressing the information of the reference image. 
Specifically, we only leverage the class token of a pretrained CLIP image encoder from the exemplar image as condition. It compresses the reference image from spatial size $224 \times 224 \times 3$ to a one-dimensional vector of dimension $1024$.

We find that this highly compressed representation tends to ignore the high-frequency details while maintaining the semantic information. It forces the network to understand the reference content and prevents the generator from directly copy-and-paste to reach the optimal results in training. For expressiveness consideration, we add several additional fully-connected (FC) layers to decode the feature and inject it into the diffusion process through cross attention.

\noindent
\textbf{Image prior.}
To further avoid the trivial solution of directly remembering the reference image, we leverage a well-trained diffusion model for initialization as a strong image prior. Specifically, we adopt a text-to-image generation model, Stable Diffusion~\cite{rombach2022high}, in consideration of two main reasons. First, it has a strong capability to generate high-quality in-the-wild images, thanks to the property that given any vector lying in its latent space will lead to a plausible image. Second, a pretrained CLIP~\cite{radford2021learning} model is utilized to extract the language information, which shares a similar representation to our adopted CLIP image embedding, making it a good initialization.




\begin{figure}[t]
\centering
\includegraphics[width=1\columnwidth]{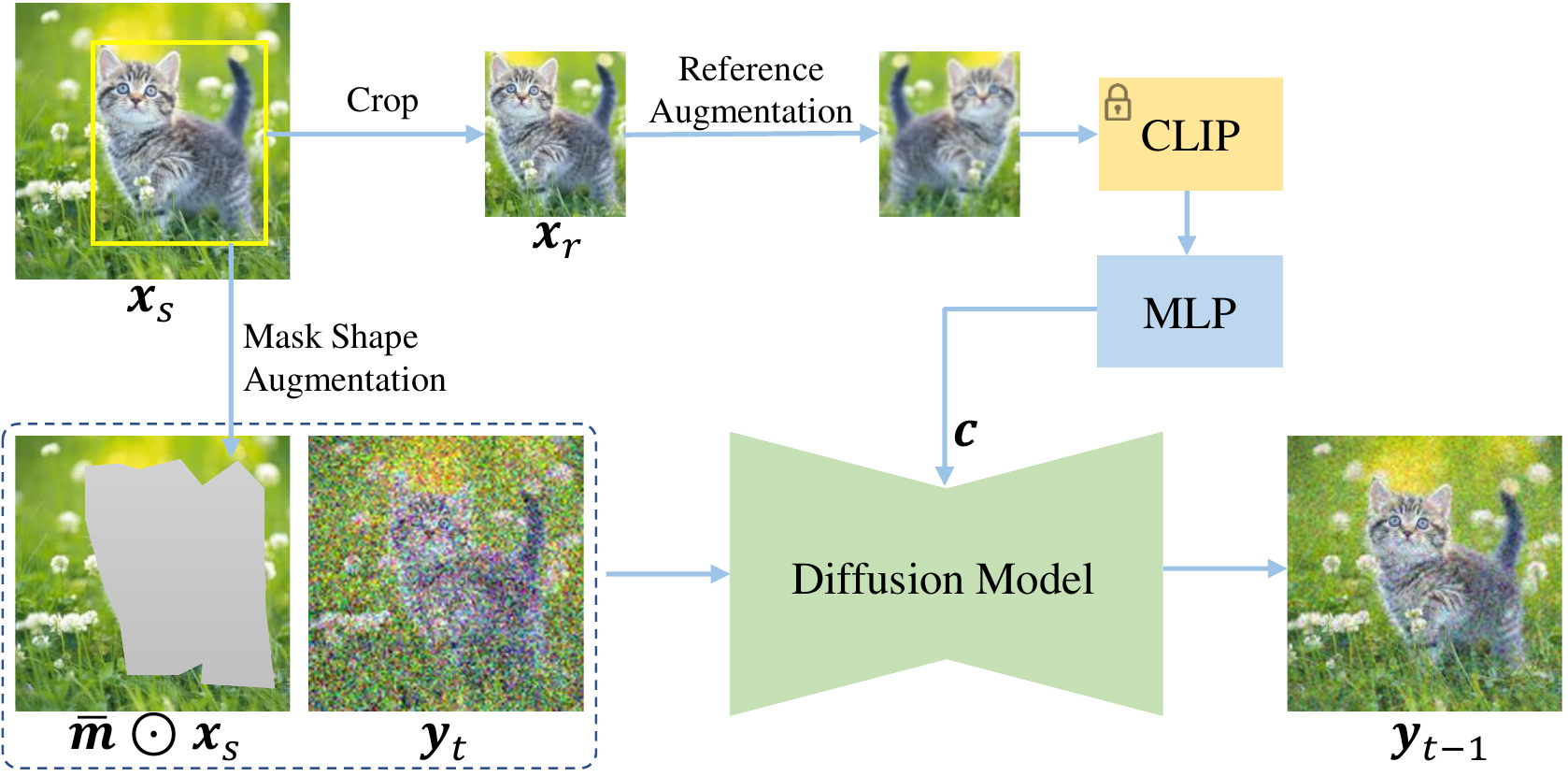}
\vspace{-0.5cm}
\caption{Our training pipeline.}
\vspace{-0.3cm}
\label{fig:pipeline}
\end{figure}

\subsubsection{Strong Augmentation}
Another potential issue of self-supervised training is the domain gap between training and testing. This train-test mismatch stems from two aspects. 

\noindent
\textbf{Reference image augmentation.}
The first mismatch is that the reference image $\mathbf{x}_r$ is derived from the source image $\mathbf{x}_s$ during training, which is barely the case for the testing scenario.
To reduce the gap, we adopt several data augmentation techniques (including flip, rotation, blur and elastic transform) on the reference image to break down the connection with the source image. We denote these data augmentation as $\mathcal{A}$.
Formally, the condition fed to the diffusion model is denoted as:
\begin{equation}
    \mathbf{c} = \text{MLP}(\text{CLIP}(\mathcal{A}(x_r))).
\end{equation}
\noindent

\noindent
\textbf{Mask shape augmentation.}
On the other hand, the mask region $\mathbf{m}$ from the bounding box ensures that the reference image contains a whole object.
As a result, the generator learns to fill an object as completely as possible.
However, this may not hold in practical scenarios. 
To address this, we generate an arbitrarily shaped mask based on the bounding box and use it in training. 
Specifically, for each edge of the bounding box, we first construct a Bessel curve to fit it, then we sample $20$ points on this curve uniformly, and randomly add $1-5$ pixel offsets to their coordinates. Finally, we connect these points with straight lines sequentially to form an arbitrarily shaped mask.
The random distortions $\mathcal{D}$ on the mask $\mathbf{m}$ break the inductive bias, reducing the gap between training and testing. \ie,
\begin{equation}
    \bar{\mathbf{m}} = \mathds{1} - \mathcal{D}(\mathbf{m}).
\end{equation}
We find these two augmentations can greatly enhance the robustness when facing different reference guidance.

\subsubsection{Control the mask shape}
Another benefit of mask shape augmentation is that it increases the control over mask shape in the inference stage. 
In practical application scenarios, a rectangle mask usually can not represent the mask area precisely. e.g. the sun umbrella in Figure~\ref{fig:teaser}. 
In some cases people would like to edit a specific region while preserving the other area as much as possible, this leads to the demand for handling irregular mask shapes. By involving these irregular masks into training, our model is able to generate photo-realistic results given various shape masks.




\subsubsection{Control the similarity degree}
To control the similarity degree between the edited area and the reference image, we find that classifier-free sampling strategy~\cite{ho2022classifier} is a powerful tool. Previous work~\cite{tang2022improved} found that the classifier-free guidance is actually the combination of both prior and posterior constraints.
\begin{equation}
\begin{split}
    &\log p(\mathbf{y}_t|\mathbf{c}) + (s-1) \log p(\mathbf{c}|\mathbf{y}_t)  \\
\propto & \log p(\mathbf{y}_t) + s(\log p(\mathbf{y}_t|\mathbf{c}) - \log p(\mathbf{y}_t)),  \\  
\end{split}
\end{equation}
where $s$ denotes the classifier-free guidance scale. It can also be regarded as the scale to control the similarity of the generated image to the reference image. A larger scale factor $s$ denotes the fusion result relies more on the conditional reference input. In our experiments, we follow the settings in ~\cite{tang2022improved} and replace $20\%$ reference conditions with a learnable vector $\mathbf{v}$ during training. This term aims to model $p(\mathbf{y}_t)$ with the help of a fixed condition input $p(\mathbf{y}_t|\mathbf{v})$. In the inference stage, each denoising step uses the modified prediction:
\begin{equation}
    \tilde{\mathbf{\epsilon}}_\theta(\mathbf{y}_t, \mathbf{c}) = 
    \mathbf{\epsilon}_\theta(\mathbf{y}_t, \mathbf{v}) + s(\mathbf{\epsilon}_\theta(\mathbf{y}_t, \mathbf{c}) - \mathbf{\epsilon}_\theta(\mathbf{y}_t, \mathbf{v})).
\end{equation}
Without causing confusion, the parameters $t$ and $\bar{\mathbf{m}} \odot \mathbf{x}_s$ are omitted here for brevity. Above all, the overall framework of our method is illustrated in Figure~\ref{fig:pipeline}.

\begin{figure*}[t]
\centering
\includegraphics[width=1\textwidth]{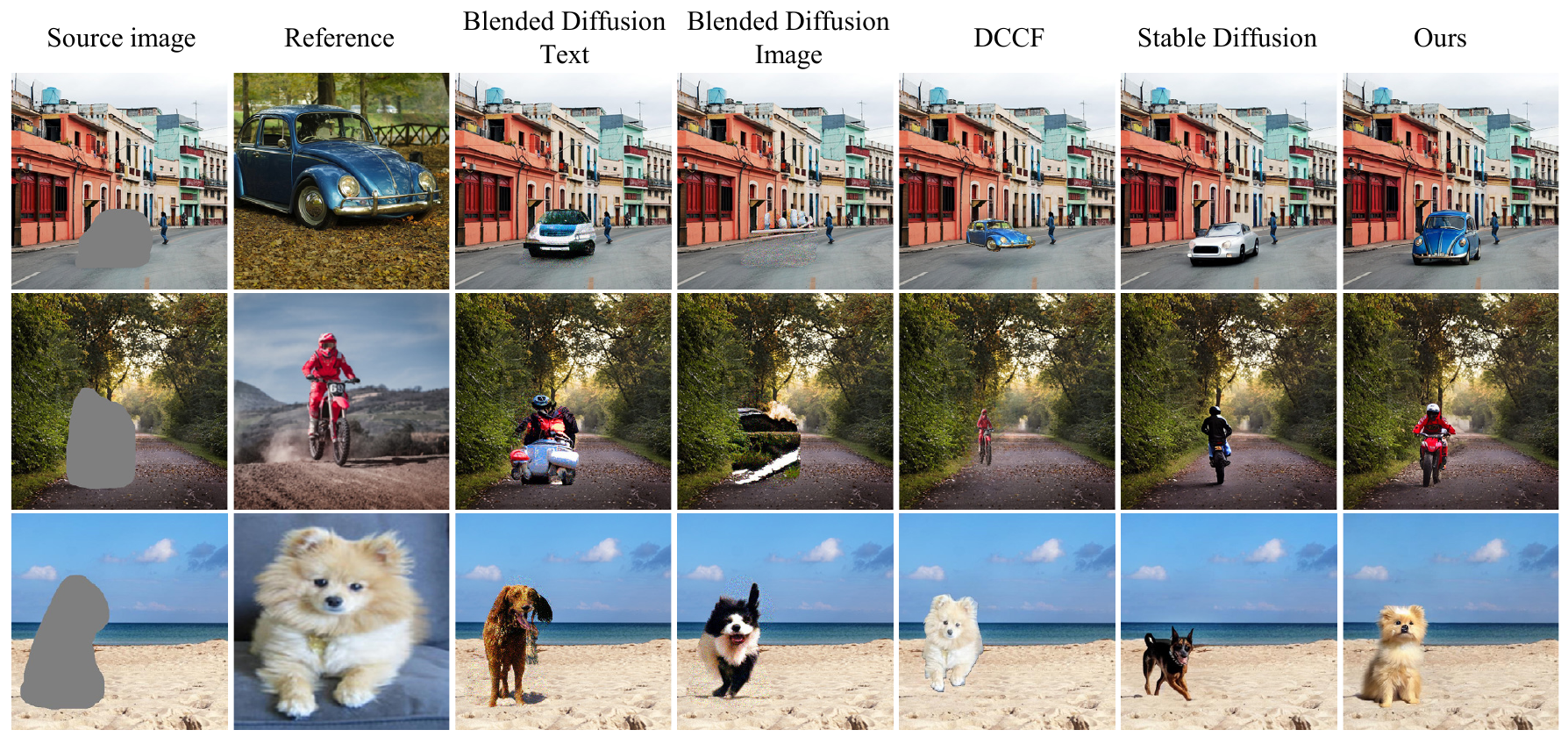}
\vspace{-0.6cm}
\caption{Qualitative comparison with other approaches. Our method can generate results that are semantically consistent with the input reference images in high perceptual quality.}
\vspace{-0.4cm}
\label{fig:other_method}
\end{figure*}

\section{Experiments}
\subsection{Implementation Details and Evaluation}
\noindent
\textbf{Implementation details.}
In order to manipulate the real-world images, first we utilize a powerful text-to-image generation model, Stable Diffusion~\cite{rombach2022high}, as initialization to provide a strong image prior. Then we select OpenImages~\cite{kuznetsova2020open} as our training dataset. It contains a total of $16$ million bounding boxes for $600$ object classes on $1.9$ million images. During training, we preprocess the image resolution to $512 \times 512$, and train our model for $40$ epochs, which takes about 7 days on 64 NVIDIA V100 GPUs. 

\noindent
\textbf{Test benchmark.}
To the best of our knowledge, no previous works target at exemplar-based semantic image editing (or semantic image composition). So we build a test benchmark for qualitative and quantitative analysis. Specifically, we manually select $3,500$ source images ($\mathbf{x}_s$) from MSCOCO~\cite{lin2014microsoft} validation set, each image contains only one bounding box ($\mathbf{m}$), and the mask region is no more than half of the whole image. Then we manually retrieve a reference image patch ($\mathbf{x}_{r}$) from MSCOCO training set. The reference image usually shares a similar semantic with mask region to ensure the combination is reasonable. We named it as COCO Exemplar-based image Editing benchmark, abbreviated as COCOEE. We will publish this benchmark, hoping to attract more follow-up works in this area.

\noindent
\textbf{Evaluation metrics.}
Our goal is to merge a reference image into a source image, while the editing region should be similar to the reference, and the fusion result should be photo-realistic. To measure these two aspects independently, we use the following three metrics to evaluate the generated images. 1) FID~\cite{heusel2017gans} score, which is widely used to evaluate generated results. We follow ~\cite{kynkaanniemi2022role} and use CLIP model to extract the feature, calculating the FID score between $3,500$ generated images and all images from COCO testing set. 2) Quality Score(QS)~\cite{gu2020giqa}, which aims to evaluate the authenticity of each single image. We take average of it to measure the overall quality of generated images. 3) CLIP score~\cite{radford2021learning},
evaluating the similarity between the edited region and the reference image. Specifically, we resize these two images to $224 \times 224$, extract the features via CLIP image encoder and calculate their cosine similarity. Higher CLIP score indicates the edited region is more similar to reference image.


\subsection{Comparisons}
Considering that no previous works aim at editing images semantically and locally based on an exemplar image, we select four related approaches as baselines to our method: 1) Blended Diffusion~\cite{avrahami2022blended}, it leverages the CLIP model to provide gradients to guide the diffusion sampling process. 
We use a text prompt ``a photo of $C$" to compute the CLIP loss, where $C$ denotes the object class from exemplar image. 2) We slightly modify the Blended Diffusion by using the reference image to calculate the CLIP loss, denoted as Blended Diffusion (image). 3) Stable Diffusion~\cite{rombach2022high}. We use the text prompt as condition to represent the reference image, and inpaint the mask region. 4) We also choose the state-of-the-art image harmonization method DCCF~\cite{xue2022dccf} as baseline. Considering it can only fuse a foreground image to background, we first use an unconditional image inpainting model LAMA~\cite{suvorov2022resolution} to inpaint the whole mask region, then extract the foreground of reference image through an additional semantic mask, and finally harmonize it into the source image.

\noindent\textbf{Qualitative analysis.}
We provide the qualitative comparison of these methods in Figure~\ref{fig:other_method}. Text-guided Blended Diffusion is able to generate objects in the desired area, but they are unrealistic and incompatible with the source image. Another text-based method Stable Diffusion can generate much realistic results, but still fail to retain the characteristics of the reference image due to the limited representation of text information. Meanwhile, the image-guided Blended Diffusion also suffers from not similar to the reference image. We argue it may caused by the gradient guidance strategy that could not preserve enough content information. Finally, the generated result from image harmonization is almost the same as the exemplar image which is very incongruous with the background. The intrinsic reason is that the appearance of exemplar image can not match the source image directly in most cases. The generative model should transform the shape, size or pose automatically to fit the source image. In the last column of Figure~\ref{fig:other_method}, our method achieves a photo-realistic result while being similar to the reference.

\noindent\textbf{Quantitative analysis.}
Table~\ref{tab:quantitative_other_methods} presents the quantitative comparison results. The image-based editing method (including Blended Diffusion (image) and DCCF) reaches a high CLIP score, demonstrating that they are able to preserve the information from condition image, while the resulting image is of poor quality. The generated result from Stable Diffusion is much more plausible according to the FID and QS. However, it can hardly incorporate the conditional information of the image. Our approach achieves the best performance on all of these three metrics, verifying that it can not only generate high-quality images but also maintain the conditional information.


%

\begin{table}[t]
\small
\centering
\setlength\tabcolsep{1.8pt}
\caption{Quantitative comparison of different methods. We evaluate the generated image quality through FID and QS, and the semantic consistency to the reference image through the CLIP score.}
\vspace{-0.2cm}
\begin{tabular}{@{}lccc@{}}
\toprule
Method   & FID ($\downarrow$) & QS ($\uparrow$)   & CLIP Score ($\uparrow$)       \\ \midrule
Blended Diffusion-Image~\cite{avrahami2022blended}   & 4.60 & 67.14 & 80.65 \\
Blended Diffusion-Text~\cite{avrahami2022blended}    & 7.52 & 55.89 & 72.62       \\
DCCF~\cite{xue2022dccf}       & 3.78 & 71.49 & 82.18 \\
Stable Diffusion~\cite{rombach2022high}    & 3.66 & 73.20 & 75.33 \\
Ours       & \textbf{3.18} & \textbf{77.80} & \textbf{84.97}            \\
\bottomrule
\end{tabular}
\label{tab:quantitative_other_methods}
\end{table}

\begin{figure*}[t]
\centering
\includegraphics[width=1\textwidth]{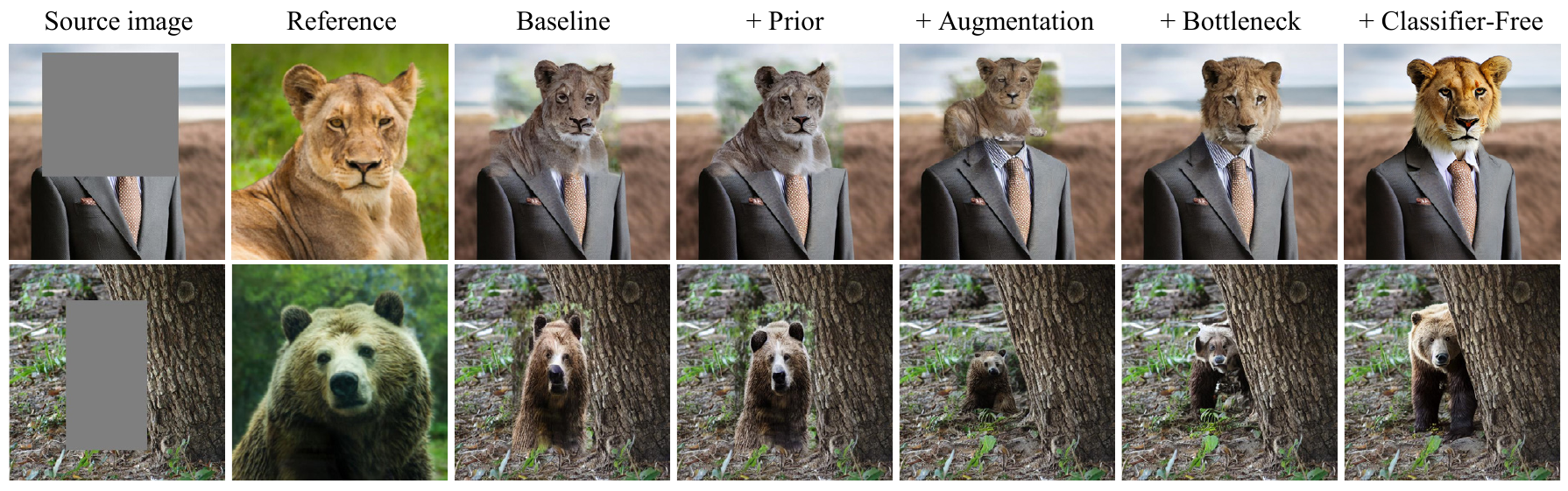}
\vspace{-0.6cm}
\caption{Visual ablation studies of individual components in our approach. We gradually eliminate the boundary artifacts through these techniques and finally achieve plausible generated results.}
\vspace{-0.3cm}
\label{fig:ablation}
\end{figure*}


\begin{figure}[t]
\centering
\includegraphics[width=1.0\linewidth]{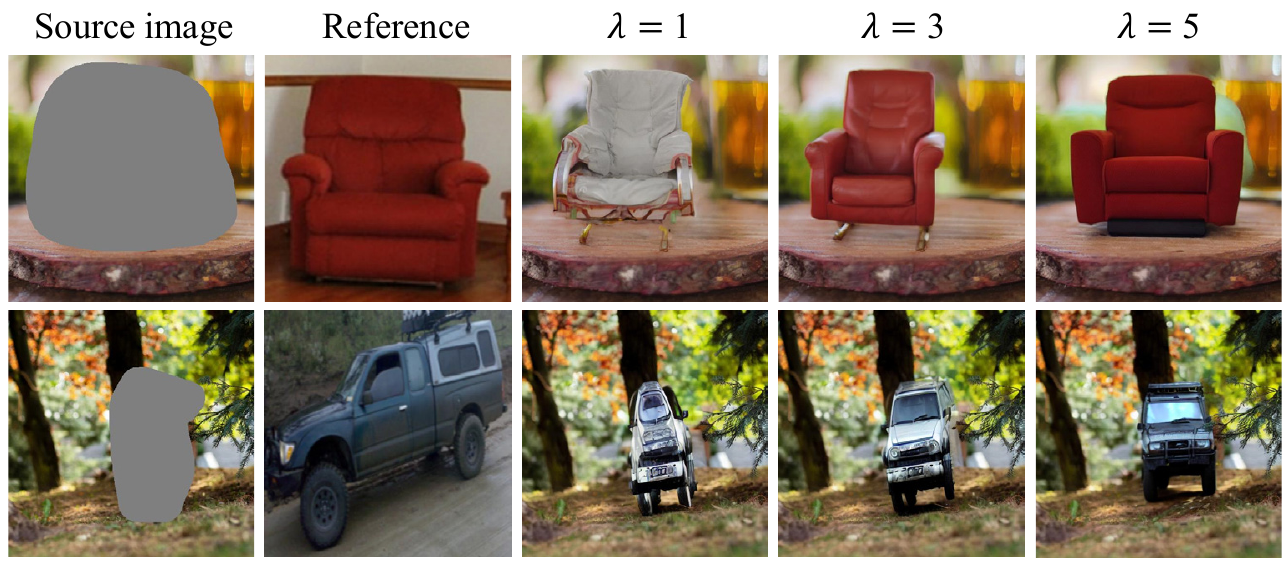}
\caption{Effect of classifier-free guidance scale $\lambda$. A larger $\lambda$ makes the generated region more similar to the reference.}
\label{fig:ablation_cf}
\vspace{-0.6cm}
\end{figure}

\noindent\textbf{User study.}
In order to obtain the user's subjective evaluation of the generated image, we conduct a user study on $50$ participants. In the study, we use $30$ groups of images, each group contains two inputs and five outputs. All these results in each group are presented side-by-side and in a random order to participants.
Participants are given unlimited time to rank the score from $1$ to $5$ ($1$ is the best, $5$ is the worst) on two perspectives independently: the image quality and the similarity to the reference image. 
We report the average ranking score in Table~\ref{tab:user_study}. Overall, the image harmonization method DCCF is most similar to reference image since it's directly copied from it. Nonetheless, users prefer our results more than others given the realistic quality of ours.

\begin{table}[t]
\small
\caption{Average ranking score of image quality and semantic consistency. $1$ is the best, $5$ is the worst. Users rated ours as the best quality, and semantic consistency is second only to the image harmonization method which copies from the exemplar image.}
\vspace{-0.2cm}
\centering
\setlength\tabcolsep{2.8pt}
\begin{tabular}{lccc}
\toprule
Method  & Quality ($\downarrow$)  & Consistency ($\downarrow$) \\
\midrule
Blended Diffusion-Image~\cite{avrahami2022blended}  & 3.83  & 3.84  \\
Blended Diffusion-Text~\cite{avrahami2022blended}  & 3.93  & 3.95   \\
DCCF~\cite{xue2022dccf}      & 3.09  &  \textbf{1.66} \\
Stable Diffusion~\cite{rombach2022high}    & 2.36 & 3.48     \\
Ours      & \textbf{1.79}  & 2.07   \\
\bottomrule
\end{tabular}
\vspace{-0.2cm}
\label{tab:user_study}
\end{table}


\begin{table}[t]
\caption{Quantitative comparison of different variants of our method. We achieve the best performance by leveraging all these techniques.}
\vspace{-0.2cm}
\small
\centering
\begin{tabular}{@{}lcccc@{}}
\toprule
Method  & FID ($\downarrow$) & QS  ($\uparrow$) & CLIP Score ($\uparrow$) \\ \midrule
Baseline    & 3.61 & 76.71 & 85.90 \\
+ Prior  & 3.40 & 77.63 & \textbf{88.79}  \\ 
+ Augmentation   & 3.44 & 76.86 & 81.68  \\
+ Bottleneck   & 3.26 & 76.62 & 81.41 \\
+ Classfier-Free    & \textbf{3.18} & \textbf{77.80} & 84.97 \\
\bottomrule
\end{tabular}
\vspace{-0.4cm}
\label{tab:quantitative_ablation}
\end{table}

\subsection{Ablation Study}
In order to achieve high-quality exemplar-based image editing, we introduce four key techniques, namely leveraging image prior, strong augmentation, information bottleneck and the classifier-free guidance. In this section, we perform five gradually changed setting to validate them:
1) We denote the naive solution in Section~\ref{method:naive} as baseline. It's directly modified from text-guided inpainting models by replacing the text to image as conditional signal. 
2) We leverage the pretrained text-to-image generation model for initialization as an image prior. 
3) To reduce the training-test gap, we adopt the strong augmentation on the reference image. 
4) To further avoid falling into the trivial solution, we highly compress the image information to increase the difficulty of reconstructing the input image during training, we denote it as the information bottleneck. 
5) At last, we use classifier-free guidance to further improve the performance. 

We show the results in Table~\ref{tab:quantitative_ablation} and Figure~\ref{fig:ablation}. The baseline solution contains obvious boundary artifacts, and makes the generated image extremely unnatural. 
By leveraging the image prior, the image quality improved according to the lower FID score. However, it still suffers from the copy-and-paste issue. 
Adding augmentations can partially alleviate it.

\begin{figure*}[t]
\centering
\includegraphics[width=1\textwidth]{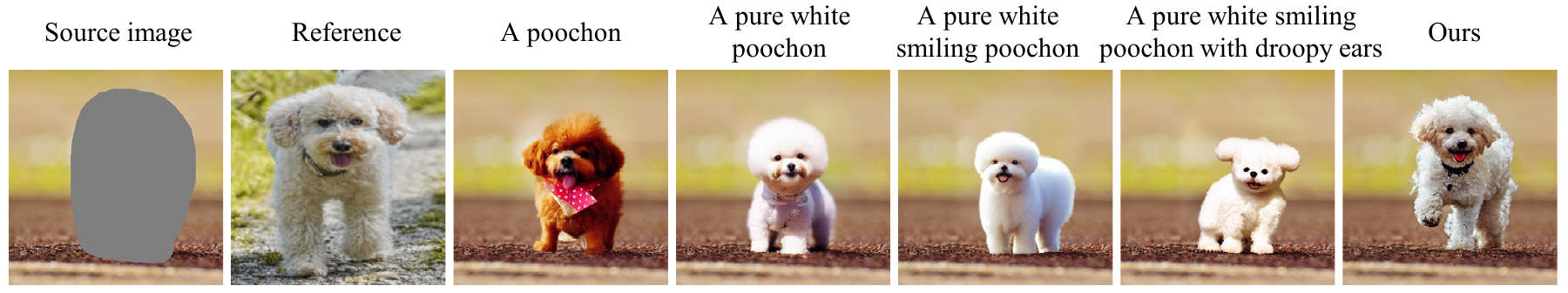}
\vspace{-0.7cm}
\caption{Comparison between progressively precise textual description and image as guidance. Using image as condition can maintain more fine-grained details.}
\vspace{-0.4cm}
\label{fig:txt_against_image}
\end{figure*}

When we further leverage the information bottleneck technique to compress the information, these boundary artifacts could be completely eliminated. Meanwhile, as the mask region should be generated instead of directly copied, the quality of this region will decrease because the difficulty of the generator increased significantly.
Finally, we add the classifier-free guidance to make the generated region more similar to the reference, it greatly boosts the overall image quality and achieves the best performance.

Meanwhile, we also investigate how the classifier-free scale affects our result. As shown in Figure~\ref{fig:ablation_cf}, as the scale $\lambda$ grows, the generated region is more and more like the reference input. In our experiment, we set $\lambda = 5$ by default.

\begin{figure}[t]
\centering
\includegraphics[width=1.0\linewidth]{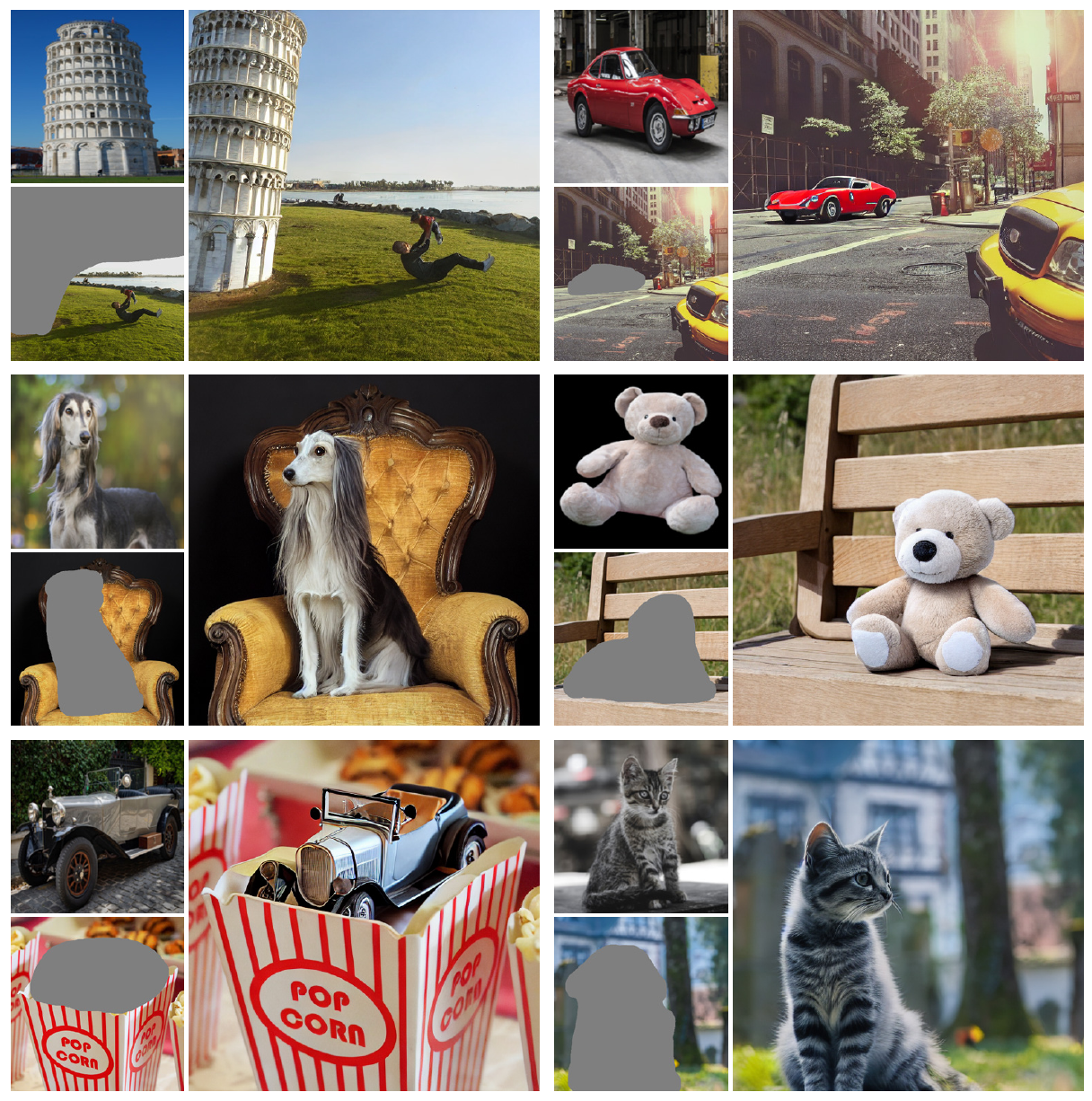}
\vspace{-0.4cm}
\caption{In-the-wild exemplar-based image editing results.}
\vspace{-0.3cm}
\label{fig:more_results}
\end{figure}

\begin{figure}[t]
\centering
\includegraphics[width=1.0\linewidth]{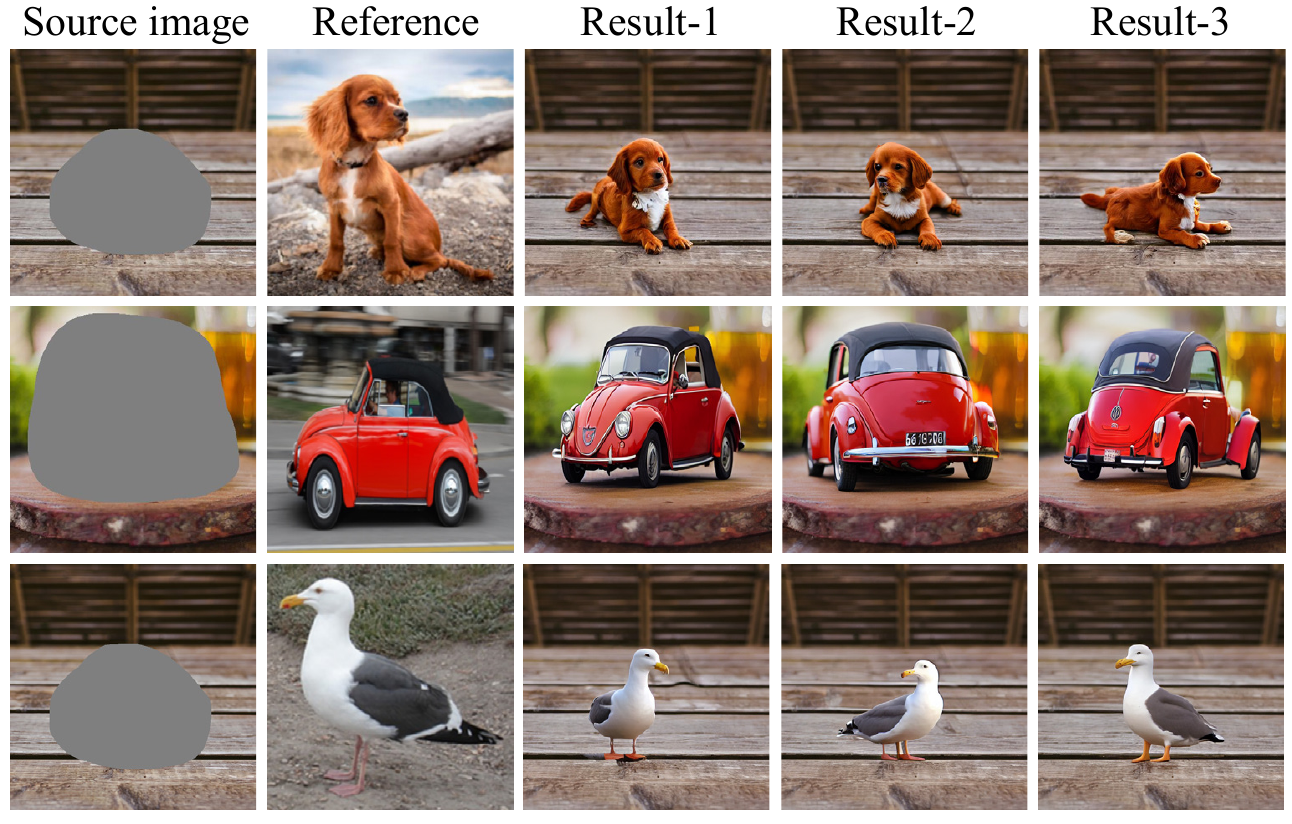}
\vspace{-0.4cm}
\caption{Our framework can synthesize realistic and diverse results from the same source image and exemplar image.}
\vspace{-0.3cm}
\label{fig:diverse_result}
\end{figure}

\subsection{From Language to Image Condition}
In Figure~\ref{fig:txt_against_image}, we provide a comparison of the controllability between language and image. From left to right, we try to inpaint the mask region with gradually detailed language description. As the language becomes more precise, the generated results indeed become more and more similar to the reference image. But it still exists a large gap with the image-guided result. Ours could maintain the fur, expression, and even the collar on the neck.

\subsection{In-the-wild Image Editing}

Benefiting from the stochasticity in the diffusion process, our method can generate multiple outputs from the same input. We show the diverse generated results in Figure~\ref{fig:diverse_result}. Although the synthesized images vary, all of them keep the key identity of the reference image. \eg, all the dogs have yellow fur, white chests and drooping ears. More selected exemplar-based image editing results are shown in Figure~\ref{fig:more_results} and appendix. 

\section{Conclusion}
\label{sec:conclusion}
In this paper, we introduce a novel image editing scenario: exemplar-based image editing, which aims to semantically alter the image content based on an exemplar image. We achieve this goal by leveraging self-supervised training based on the diffusion model. The naive approach causes the boundary artifacts issue, we carefully analyze it and solve it by proposing a group of techniques. Our algorithm enables the user to precisely control the editing, and achieves an impressive performance on in-the-wild images. We hope this work will serve as a solid baseline and help support future research in exemplar-based image editing area.
{\small
\bibliographystyle{ieee_fullname}
\bibliography{egbib}

\begin{thebibliography}{10}\itemsep=-1pt

\bibitem{abdal2022clip2stylegan}
Rameen Abdal, Peihao Zhu, John Femiani, Niloy Mitra, and Peter Wonka.
\newblock Clip2stylegan: Unsupervised extraction of stylegan edit directions.
\newblock In {\em ACM SIGGRAPH 2022 Conference Proceedings}, pages 1--9, 2022.

\bibitem{alaluf2022hyperstyle}
Yuval Alaluf, Omer Tov, Ron Mokady, Rinon Gal, and Amit Bermano.
\newblock Hyperstyle: Stylegan inversion with hypernetworks for real image
  editing.
\newblock In {\em Proceedings of the IEEE/CVF Conference on Computer Vision and
  Pattern Recognition}, pages 18511--18521, 2022.

\bibitem{avrahami2022blended}
Omri Avrahami, Dani Lischinski, and Ohad Fried.
\newblock Blended diffusion for text-driven editing of natural images.
\newblock In {\em Proceedings of the IEEE/CVF Conference on Computer Vision and
  Pattern Recognition}, pages 18208--18218, 2022.

\bibitem{bau2021paint}
David Bau, Alex Andonian, Audrey Cui, YeonHwan Park, Ali Jahanian, Aude Oliva,
  and Antonio Torralba.
\newblock Paint by word.
\newblock {\em arXiv preprint arXiv:2103.10951}, 2021.

\bibitem{bau2020semantic}
David Bau, Hendrik Strobelt, William Peebles, Jonas Wulff, Bolei Zhou, Jun-Yan
  Zhu, and Antonio Torralba.
\newblock Semantic photo manipulation with a generative image prior.
\newblock {\em arXiv preprint arXiv:2005.07727}, 2020.

\bibitem{info11020125}
Alexander Buslaev, Vladimir~I. Iglovikov, Eugene Khvedchenya, Alex Parinov,
  Mikhail Druzhinin, and Alexandr~A. Kalinin.
\newblock Albumentations: Fast and flexible image augmentations.
\newblock {\em Information}, 11(2), 2020.

\bibitem{chen2019toward}
Bor-Chun Chen and Andrew Kae.
\newblock Toward realistic image compositing with adversarial learning.
\newblock In {\em Proceedings of the IEEE/CVF Conference on Computer Vision and
  Pattern Recognition}, pages 8415--8424, 2019.

\bibitem{chen2018deep}
Yu-Sheng Chen, Yu-Ching Wang, Man-Hsin Kao, and Yung-Yu Chuang.
\newblock Deep photo enhancer: Unpaired learning for image enhancement from
  photographs with gans.
\newblock In {\em Proceedings of the IEEE Conference on Computer Vision and
  Pattern Recognition}, pages 6306--6314, 2018.

\bibitem{cohen2006color}
Daniel Cohen-Or, Olga Sorkine, Ran Gal, Tommer Leyvand, and Ying-Qing Xu.
\newblock Color harmonization.
\newblock In {\em ACM SIGGRAPH 2006 Papers}, pages 624--630. 2006.

\bibitem{cong2020dovenet}
Wenyan Cong, Jianfu Zhang, Li Niu, Liu Liu, Zhixin Ling, Weiyuan Li, and Liqing
  Zhang.
\newblock Dovenet: Deep image harmonization via domain verification.
\newblock In {\em Proceedings of the IEEE/CVF Conference on Computer Vision and
  Pattern Recognition}, pages 8394--8403, 2020.

\bibitem{cun2020improving}
Xiaodong Cun and Chi-Man Pun.
\newblock Improving the harmony of the composite image by spatial-separated
  attention module.
\newblock {\em IEEE Transactions on Image Processing}, 29:4759--4771, 2020.

\bibitem{deng2018aesthetic}
Yubin Deng, Chen~Change Loy, and Xiaoou Tang.
\newblock Aesthetic-driven image enhancement by adversarial learning.
\newblock In {\em Proceedings of the 26th ACM international conference on
  Multimedia}, pages 870--878, 2018.

\bibitem{gafni2022make}
Oran Gafni, Adam Polyak, Oron Ashual, Shelly Sheynin, Devi Parikh, and Yaniv
  Taigman.
\newblock Make-a-scene: Scene-based text-to-image generation with human priors.
\newblock {\em arXiv preprint arXiv:2203.13131}, 2022.

\bibitem{gal2022stylegan}
Rinon Gal, Or Patashnik, Haggai Maron, Amit~H Bermano, Gal Chechik, and Daniel
  Cohen-Or.
\newblock Stylegan-nada: Clip-guided domain adaptation of image generators.
\newblock {\em ACM Transactions on Graphics (TOG)}, 41(4):1--13, 2022.

\bibitem{gao2021high}
Yue Gao, Fangyun Wei, Jianmin Bao, Shuyang Gu, Dong Chen, Fang Wen, and Zhouhui
  Lian.
\newblock High-fidelity and arbitrary face editing.
\newblock In {\em Proceedings of the IEEE/CVF conference on computer vision and
  pattern recognition}, pages 16115--16124, 2021.

\bibitem{goodfellow2020generative}
Ian Goodfellow, Jean Pouget-Abadie, Mehdi Mirza, Bing Xu, David Warde-Farley,
  Sherjil Ozair, Aaron Courville, and Yoshua Bengio.
\newblock Generative adversarial networks.
\newblock {\em Communications of the ACM}, 63(11):139--144, 2020.

\bibitem{gu2020giqa}
Shuyang Gu, Jianmin Bao, Dong Chen, and Fang Wen.
\newblock Giqa: Generated image quality assessment.
\newblock In {\em European conference on computer vision}, pages 369--385.
  Springer, 2020.

\bibitem{gu2019mask}
Shuyang Gu, Jianmin Bao, Hao Yang, Dong Chen, Fang Wen, and Lu Yuan.
\newblock Mask-guided portrait editing with conditional gans.
\newblock In {\em Proceedings of the IEEE/CVF conference on computer vision and
  pattern recognition}, pages 3436--3445, 2019.

\bibitem{gu2022vector}
Shuyang Gu, Dong Chen, Jianmin Bao, Fang Wen, Bo Zhang, Dongdong Chen, Lu Yuan,
  and Baining Guo.
\newblock Vector quantized diffusion model for text-to-image synthesis.
\newblock In {\em Proceedings of the IEEE/CVF Conference on Computer Vision and
  Pattern Recognition}, pages 10696--10706, 2022.

\bibitem{guo2021image}
Zonghui Guo, Dongsheng Guo, Haiyong Zheng, Zhaorui Gu, Bing Zheng, and Junyu
  Dong.
\newblock Image harmonization with transformer.
\newblock In {\em Proceedings of the IEEE/CVF International Conference on
  Computer Vision}, pages 14870--14879, 2021.

\bibitem{guo2021intrinsic}
Zonghui Guo, Haiyong Zheng, Yufeng Jiang, Zhaorui Gu, and Bing Zheng.
\newblock Intrinsic image harmonization.
\newblock In {\em Proceedings of the IEEE/CVF Conference on Computer Vision and
  Pattern Recognition}, pages 16367--16376, 2021.

\bibitem{hertz2022prompt}
Amir Hertz, Ron Mokady, Jay Tenenbaum, Kfir Aberman, Yael Pritch, and Daniel
  Cohen-Or.
\newblock Prompt-to-prompt image editing with cross attention control.
\newblock {\em arXiv preprint arXiv:2208.01626}, 2022.

\bibitem{heusel2017gans}
Martin Heusel, Hubert Ramsauer, Thomas Unterthiner, Bernhard Nessler, and Sepp
  Hochreiter.
\newblock Gans trained by a two time-scale update rule converge to a local nash
  equilibrium.
\newblock {\em Advances in neural information processing systems}, 30, 2017.

\bibitem{ho2020denoising}
Jonathan Ho, Ajay Jain, and Pieter Abbeel.
\newblock Denoising diffusion probabilistic models.
\newblock {\em Advances in Neural Information Processing Systems},
  33:6840--6851, 2020.

\bibitem{ho2022classifier}
Jonathan Ho and Tim Salimans.
\newblock Classifier-free diffusion guidance.
\newblock {\em arXiv preprint arXiv:2207.12598}, 2022.

\bibitem{hou2022guidedstyle}
Xianxu Hou, Xiaokang Zhang, Hanbang Liang, Linlin Shen, Zhihui Lai, and Jun
  Wan.
\newblock Guidedstyle: Attribute knowledge guided style manipulation for
  semantic face editing.
\newblock {\em Neural Networks}, 145:209--220, 2022.

\bibitem{jia2006drag}
Jiaya Jia, Jian Sun, Chi-Keung Tang, and Heung-Yeung Shum.
\newblock Drag-and-drop pasting.
\newblock {\em ACM Transactions on graphics (TOG)}, 25(3):631--637, 2006.

\bibitem{karras2019style}
Tero Karras, Samuli Laine, and Timo Aila.
\newblock A style-based generator architecture for generative adversarial
  networks.
\newblock In {\em Proceedings of the IEEE/CVF conference on computer vision and
  pattern recognition}, pages 4401--4410, 2019.

\bibitem{karras2020analyzing}
Tero Karras, Samuli Laine, Miika Aittala, Janne Hellsten, Jaakko Lehtinen, and
  Timo Aila.
\newblock Analyzing and improving the image quality of stylegan.
\newblock In {\em Proceedings of the IEEE/CVF conference on computer vision and
  pattern recognition}, pages 8110--8119, 2020.

\bibitem{kawar2022imagic}
Bahjat Kawar, Shiran Zada, Oran Lang, Omer Tov, Huiwen Chang, Tali Dekel, Inbar
  Mosseri, and Michal Irani.
\newblock Imagic: Text-based real image editing with diffusion models.
\newblock {\em arXiv preprint arXiv:2210.09276}, 2022.

\bibitem{kim2022diffusionclip}
Gwanghyun Kim, Taesung Kwon, and Jong~Chul Ye.
\newblock Diffusionclip: Text-guided diffusion models for robust image
  manipulation.
\newblock In {\em Proceedings of the IEEE/CVF Conference on Computer Vision and
  Pattern Recognition}, pages 2426--2435, 2022.

\bibitem{kuznetsova2020open}
Alina Kuznetsova, Hassan Rom, Neil Alldrin, Jasper Uijlings, Ivan Krasin, Jordi
  Pont-Tuset, Shahab Kamali, Stefan Popov, Matteo Malloci, Alexander
  Kolesnikov, et~al.
\newblock The open images dataset v4.
\newblock {\em International Journal of Computer Vision}, 128(7):1956--1981,
  2020.

\bibitem{kynkaanniemi2022role}
Tuomas Kynk{\"a}{\"a}nniemi, Tero Karras, Miika Aittala, Timo Aila, and Jaakko
  Lehtinen.
\newblock The role of imagenet classes in fr$\backslash$'echet inception
  distance.
\newblock {\em arXiv preprint arXiv:2203.06026}, 2022.

\bibitem{li2019faceshifter}
Lingzhi Li, Jianmin Bao, Hao Yang, Dong Chen, and Fang Wen.
\newblock Faceshifter: Towards high fidelity and occlusion aware face swapping.
\newblock {\em arXiv preprint arXiv:1912.13457}, 2019.

\bibitem{lin2014microsoft}
Tsung-Yi Lin, Michael Maire, Serge Belongie, James Hays, Pietro Perona, Deva
  Ramanan, Piotr Doll{\'a}r, and C~Lawrence Zitnick.
\newblock Microsoft coco: Common objects in context.
\newblock In {\em European conference on computer vision}, pages 740--755.
  Springer, 2014.

\bibitem{ling2021editgan}
Huan Ling, Karsten Kreis, Daiqing Li, Seung~Wook Kim, Antonio Torralba, and
  Sanja Fidler.
\newblock Editgan: High-precision semantic image editing.
\newblock {\em Advances in Neural Information Processing Systems},
  34:16331--16345, 2021.

\bibitem{liu2021generative}
Ming-Yu Liu, Xun Huang, Jiahui Yu, Ting-Chun Wang, and Arun Mallya.
\newblock Generative adversarial networks for image and video synthesis:
  Algorithms and applications.
\newblock {\em Proceedings of the IEEE}, 109(5):839--862, 2021.

\bibitem{liu2021more}
Xihui Liu, Dong~Huk Park, Samaneh Azadi, Gong Zhang, Arman Chopikyan, Yuxiao
  Hu, Humphrey Shi, Anna Rohrbach, and Trevor Darrell.
\newblock More control for free! image synthesis with semantic diffusion
  guidance.
\newblock {\em arXiv preprint arXiv:2112.05744}, 2021.

\bibitem{loshchilov2017decoupled}
Ilya Loshchilov and Frank Hutter.
\newblock Decoupled weight decay regularization.
\newblock {\em arXiv preprint arXiv:1711.05101}, 2017.

\bibitem{nichol2021glide}
Alex Nichol, Prafulla Dhariwal, Aditya Ramesh, Pranav Shyam, Pamela Mishkin,
  Bob McGrew, Ilya Sutskever, and Mark Chen.
\newblock Glide: Towards photorealistic image generation and editing with
  text-guided diffusion models.
\newblock {\em arXiv preprint arXiv:2112.10741}, 2021.

\bibitem{GLIDE}
Alex Nichol, Prafulla Dhariwal, Aditya Ramesh, Pranav Shyam, Pamela Mishkin,
  Bob McGrew, Ilya Sutskever, and Mark Chen.
\newblock Glide: Towards photorealistic image generation and editing with
  text-guided diffusion models.
\newblock {\em arXiv preprint arXiv:2112.10741}, 2021.

\bibitem{niu2021making}
Li Niu, Wenyan Cong, Liu Liu, Yan Hong, Bo Zhang, Jing Liang, and Liqing Zhang.
\newblock Making images real again: A comprehensive survey on deep image
  composition.
\newblock {\em arXiv preprint arXiv:2106.14490}, 2021.

\bibitem{patashnik2021styleclip}
Or Patashnik, Zongze Wu, Eli Shechtman, Daniel Cohen-Or, and Dani Lischinski.
\newblock Styleclip: Text-driven manipulation of stylegan imagery.
\newblock In {\em Proceedings of the IEEE/CVF International Conference on
  Computer Vision}, pages 2085--2094, 2021.

\bibitem{perez2003poisson}
Patrick P{\'e}rez, Michel Gangnet, and Andrew Blake.
\newblock Poisson image editing.
\newblock In {\em ACM SIGGRAPH 2003 Papers}, pages 313--318. 2003.

\bibitem{radford2021learning}
Alec Radford, Jong~Wook Kim, Chris Hallacy, Aditya Ramesh, Gabriel Goh,
  Sandhini Agarwal, Girish Sastry, Amanda Askell, Pamela Mishkin, Jack Clark,
  et~al.
\newblock Learning transferable visual models from natural language
  supervision.
\newblock In {\em International Conference on Machine Learning}, pages
  8748--8763. PMLR, 2021.

\bibitem{ramesh2022hierarchical}
Aditya Ramesh, Prafulla Dhariwal, Alex Nichol, Casey Chu, and Mark Chen.
\newblock Hierarchical text-conditional image generation with clip latents.
\newblock {\em arXiv preprint arXiv:2204.06125}, 2022.

\bibitem{ramesh2021zero}
Aditya Ramesh, Mikhail Pavlov, Gabriel Goh, Scott Gray, Chelsea Voss, Alec
  Radford, Mark Chen, and Ilya Sutskever.
\newblock Zero-shot text-to-image generation.
\newblock In {\em International Conference on Machine Learning}, pages
  8821--8831. PMLR, 2021.

\bibitem{reinhard2001color}
Erik Reinhard, Michael Adhikhmin, Bruce Gooch, and Peter Shirley.
\newblock Color transfer between images.
\newblock {\em IEEE Computer graphics and applications}, 21(5):34--41, 2001.

\bibitem{richardson2021encoding}
Elad Richardson, Yuval Alaluf, Or Patashnik, Yotam Nitzan, Yaniv Azar, Stav
  Shapiro, and Daniel Cohen-Or.
\newblock Encoding in style: a stylegan encoder for image-to-image translation.
\newblock In {\em Proceedings of the IEEE/CVF conference on computer vision and
  pattern recognition}, pages 2287--2296, 2021.

\bibitem{roich2022pivotal}
Daniel Roich, Ron Mokady, Amit~H Bermano, and Daniel Cohen-Or.
\newblock Pivotal tuning for latent-based editing of real images.
\newblock {\em ACM Transactions on Graphics (TOG)}, 42(1):1--13, 2022.

\bibitem{rombach2021highresolution}
Robin Rombach, Andreas Blattmann, Dominik Lorenz, Patrick Esser, and Björn
  Ommer.
\newblock High-resolution image synthesis with latent diffusion models, 2021.

\bibitem{rombach2022high}
Robin Rombach, Andreas Blattmann, Dominik Lorenz, Patrick Esser, and Bj{\"o}rn
  Ommer.
\newblock High-resolution image synthesis with latent diffusion models.
\newblock In {\em Proceedings of the IEEE/CVF Conference on Computer Vision and
  Pattern Recognition}, pages 10684--10695, 2022.

\bibitem{ruiz2022dreambooth}
Nataniel Ruiz, Yuanzhen Li, Varun Jampani, Yael Pritch, Michael Rubinstein, and
  Kfir Aberman.
\newblock Dreambooth: Fine tuning text-to-image diffusion models for
  subject-driven generation.
\newblock {\em arXiv preprint arXiv:2208.12242}, 2022.

\bibitem{saharia2022palette}
Chitwan Saharia, William Chan, Huiwen Chang, Chris Lee, Jonathan Ho, Tim
  Salimans, David Fleet, and Mohammad Norouzi.
\newblock Palette: Image-to-image diffusion models.
\newblock In {\em ACM SIGGRAPH 2022 Conference Proceedings}, pages 1--10, 2022.

\bibitem{saharia2022photorealistic}
Chitwan Saharia, William Chan, Saurabh Saxena, Lala Li, Jay Whang, Emily
  Denton, Seyed Kamyar~Seyed Ghasemipour, Burcu~Karagol Ayan, S~Sara Mahdavi,
  Rapha~Gontijo Lopes, et~al.
\newblock Photorealistic text-to-image diffusion models with deep language
  understanding.
\newblock {\em arXiv preprint arXiv:2205.11487}, 2022.

\bibitem{shen2020interpreting}
Yujun Shen, Jinjin Gu, Xiaoou Tang, and Bolei Zhou.
\newblock Interpreting the latent space of gans for semantic face editing.
\newblock In {\em Proceedings of the IEEE/CVF conference on computer vision and
  pattern recognition}, pages 9243--9252, 2020.

\bibitem{shen2018faceid}
Yujun Shen, Ping Luo, Junjie Yan, Xiaogang Wang, and Xiaoou Tang.
\newblock Faceid-gan: Learning a symmetry three-player gan for
  identity-preserving face synthesis.
\newblock In {\em Proceedings of the IEEE conference on computer vision and
  pattern recognition}, pages 821--830, 2018.

\bibitem{shen2021closed}
Yujun Shen and Bolei Zhou.
\newblock Closed-form factorization of latent semantics in gans.
\newblock In {\em Proceedings of the IEEE/CVF Conference on Computer Vision and
  Pattern Recognition}, pages 1532--1540, 2021.

\bibitem{song2020score}
Yang Song, Jascha Sohl-Dickstein, Diederik~P Kingma, Abhishek Kumar, Stefano
  Ermon, and Ben Poole.
\newblock Score-based generative modeling through stochastic differential
  equations.
\newblock {\em arXiv preprint arXiv:2011.13456}, 2020.

\bibitem{sunkavalli2010multi}
Kalyan Sunkavalli, Micah~K Johnson, Wojciech Matusik, and Hanspeter Pfister.
\newblock Multi-scale image harmonization.
\newblock {\em ACM Transactions on Graphics (TOG)}, 29(4):1--10, 2010.

\bibitem{suvorov2022resolution}
Roman Suvorov, Elizaveta Logacheva, Anton Mashikhin, Anastasia Remizova,
  Arsenii Ashukha, Aleksei Silvestrov, Naejin Kong, Harshith Goka, Kiwoong
  Park, and Victor Lempitsky.
\newblock Resolution-robust large mask inpainting with fourier convolutions.
\newblock In {\em Proceedings of the IEEE/CVF Winter Conference on Applications
  of Computer Vision}, pages 2149--2159, 2022.

\bibitem{tang2022improved}
Zhicong Tang, Shuyang Gu, Jianmin Bao, Dong Chen, and Fang Wen.
\newblock Improved vector quantized diffusion models.
\newblock {\em arXiv preprint arXiv:2205.16007}, 2022.

\bibitem{tao2010error}
Michael~W Tao, Micah~K Johnson, and Sylvain Paris.
\newblock Error-tolerant image compositing.
\newblock In {\em European Conference on Computer Vision}, pages 31--44.
  Springer, 2010.

\bibitem{tsai2017deep}
Yi-Hsuan Tsai, Xiaohui Shen, Zhe Lin, Kalyan Sunkavalli, Xin Lu, and Ming-Hsuan
  Yang.
\newblock Deep image harmonization.
\newblock In {\em Proceedings of the IEEE Conference on Computer Vision and
  Pattern Recognition}, pages 3789--3797, 2017.

\bibitem{wang2022pretraining}
Tengfei Wang, Ting Zhang, Bo Zhang, Hao Ouyang, Dong Chen, Qifeng Chen, and
  Fang Wen.
\newblock Pretraining is all you need for image-to-image translation.
\newblock {\em arXiv preprint arXiv:2205.12952}, 2022.

\bibitem{xia2021tedigan}
Weihao Xia, Yujiu Yang, Jing-Hao Xue, and Baoyuan Wu.
\newblock Tedigan: Text-guided diverse face image generation and manipulation.
\newblock In {\em Proceedings of the IEEE/CVF conference on computer vision and
  pattern recognition}, pages 2256--2265, 2021.

\bibitem{xue2022dccf}
Ben Xue, Shenghui Ran, Quan Chen, Rongfei Jia, Binqiang Zhao, and Xing Tang.
\newblock Dccf: Deep comprehensible color filter learning framework for
  high-resolution image harmonization.
\newblock {\em arXiv preprint arXiv:2207.04788}, 2022.

\bibitem{yu2018generative}
Jiahui Yu, Zhe Lin, Jimei Yang, Xiaohui Shen, Xin Lu, and Thomas~S Huang.
\newblock Generative image inpainting with contextual attention.
\newblock In {\em Proceedings of the IEEE conference on computer vision and
  pattern recognition}, pages 5505--5514, 2018.

\bibitem{yu2022scaling}
Jiahui Yu, Yuanzhong Xu, Jing~Yu Koh, Thang Luong, Gunjan Baid, Zirui Wang,
  Vijay Vasudevan, Alexander Ku, Yinfei Yang, Burcu~Karagol Ayan, et~al.
\newblock Scaling autoregressive models for content-rich text-to-image
  generation.
\newblock {\em arXiv preprint arXiv:2206.10789}, 2022.

\bibitem{zhang2022styleswin}
Bowen Zhang, Shuyang Gu, Bo Zhang, Jianmin Bao, Dong Chen, Fang Wen, Yong Wang,
  and Baining Guo.
\newblock Styleswin: Transformer-based gan for high-resolution image
  generation.
\newblock In {\em Proceedings of the IEEE/CVF Conference on Computer Vision and
  Pattern Recognition}, pages 11304--11314, 2022.

\bibitem{zhang2019deep}
Bo Zhang, Mingming He, Jing Liao, Pedro~V Sander, Lu Yuan, Amine Bermak, and
  Dong Chen.
\newblock Deep exemplar-based video colorization.
\newblock In {\em Proceedings of the IEEE/CVF Conference on Computer Vision and
  Pattern Recognition}, pages 8052--8061, 2019.

\bibitem{zhang2021deep}
He Zhang, Jianming Zhang, Federico Perazzi, Zhe Lin, and Vishal~M Patel.
\newblock Deep image compositing.
\newblock In {\em Proceedings of the IEEE/CVF Winter Conference on Applications
  of Computer Vision}, pages 365--374, 2021.

\bibitem{zhang2020cross}
Pan Zhang, Bo Zhang, Dong Chen, Lu Yuan, and Fang Wen.
\newblock Cross-domain correspondence learning for exemplar-based image
  translation.
\newblock In {\em Proceedings of the IEEE/CVF Conference on Computer Vision and
  Pattern Recognition}, pages 5143--5153, 2020.

\bibitem{zhang2016colorful}
Richard Zhang, Phillip Isola, and Alexei~A Efros.
\newblock Colorful image colorization.
\newblock In {\em European conference on computer vision}, pages 649--666.
  Springer, 2016.

\bibitem{zhou2021cocosnet}
Xingran Zhou, Bo Zhang, Ting Zhang, Pan Zhang, Jianmin Bao, Dong Chen, Zhongfei
  Zhang, and Fang Wen.
\newblock Cocosnet v2: Full-resolution correspondence learning for image
  translation.
\newblock In {\em Proceedings of the IEEE/CVF Conference on Computer Vision and
  Pattern Recognition}, pages 11465--11475, 2021.

\end{thebibliography}
}

\appendix
\onecolumn{
\begin{appendices}
\section{Additional results}
In this section, we provide additional results of our method in different application scenarios. 
Fig.~\ref{fig:supp_2} demonstrates the ability of our method in editing any region of real images. Our method is able to understand the objects in the reference images and generate corresponding objects in the edited region. The generated objects are highly in harmony with the source images.
In Fig.~\ref{fig:supp_1}, we show results of the same object with different source images, which demonstrates the robustness of our method. 

\begin{figure*}[h]
\centering
\vspace{0cm}
\includegraphics[width=0.95\textwidth]{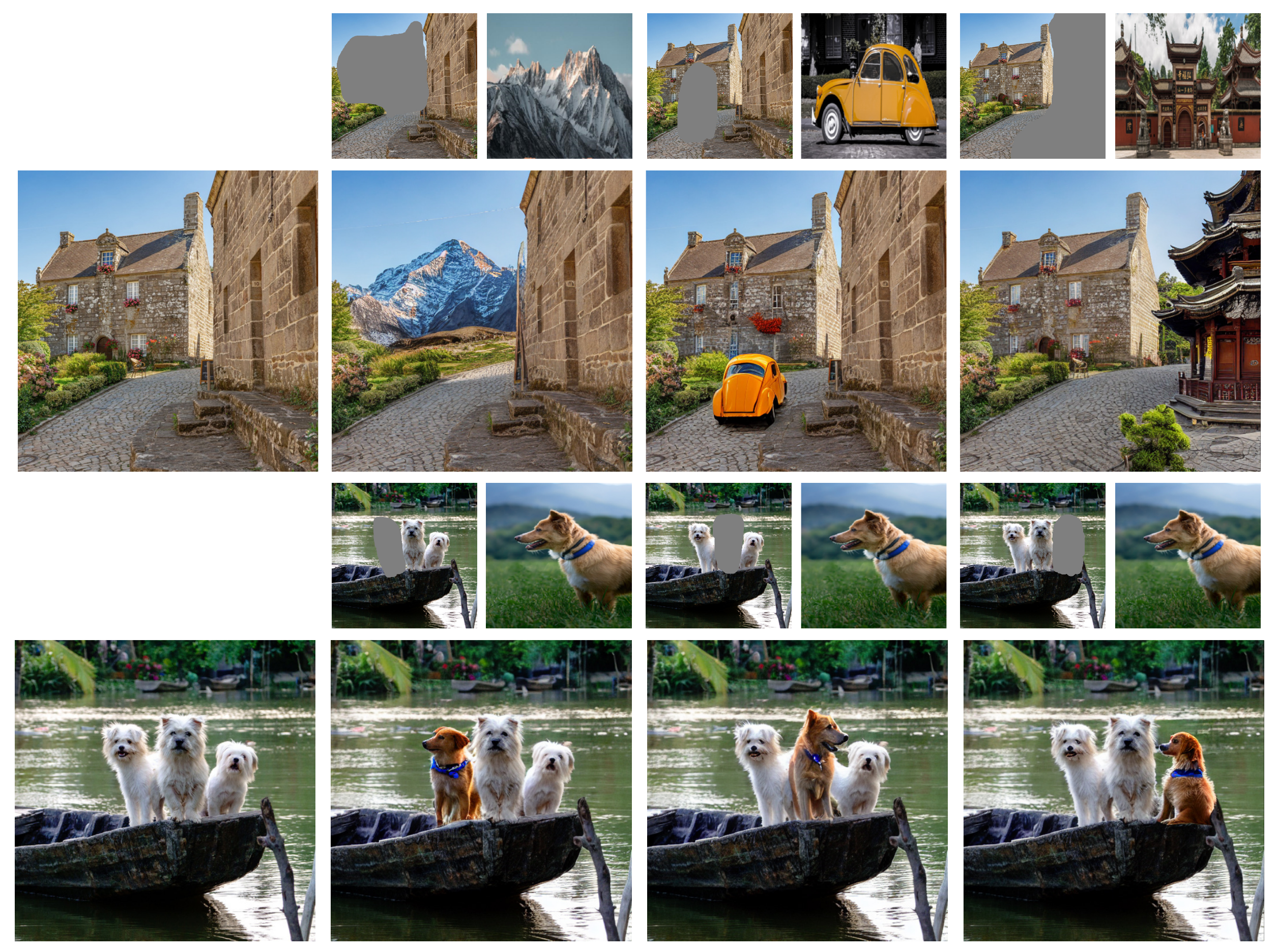}
\caption{Our method enables the user to edit different regions in the same source images.}
\label{fig:supp_2}
\vspace{-0.3cm}
\end{figure*}

\begin{figure*}[t]
\centering
\includegraphics[width=1.0\textwidth]{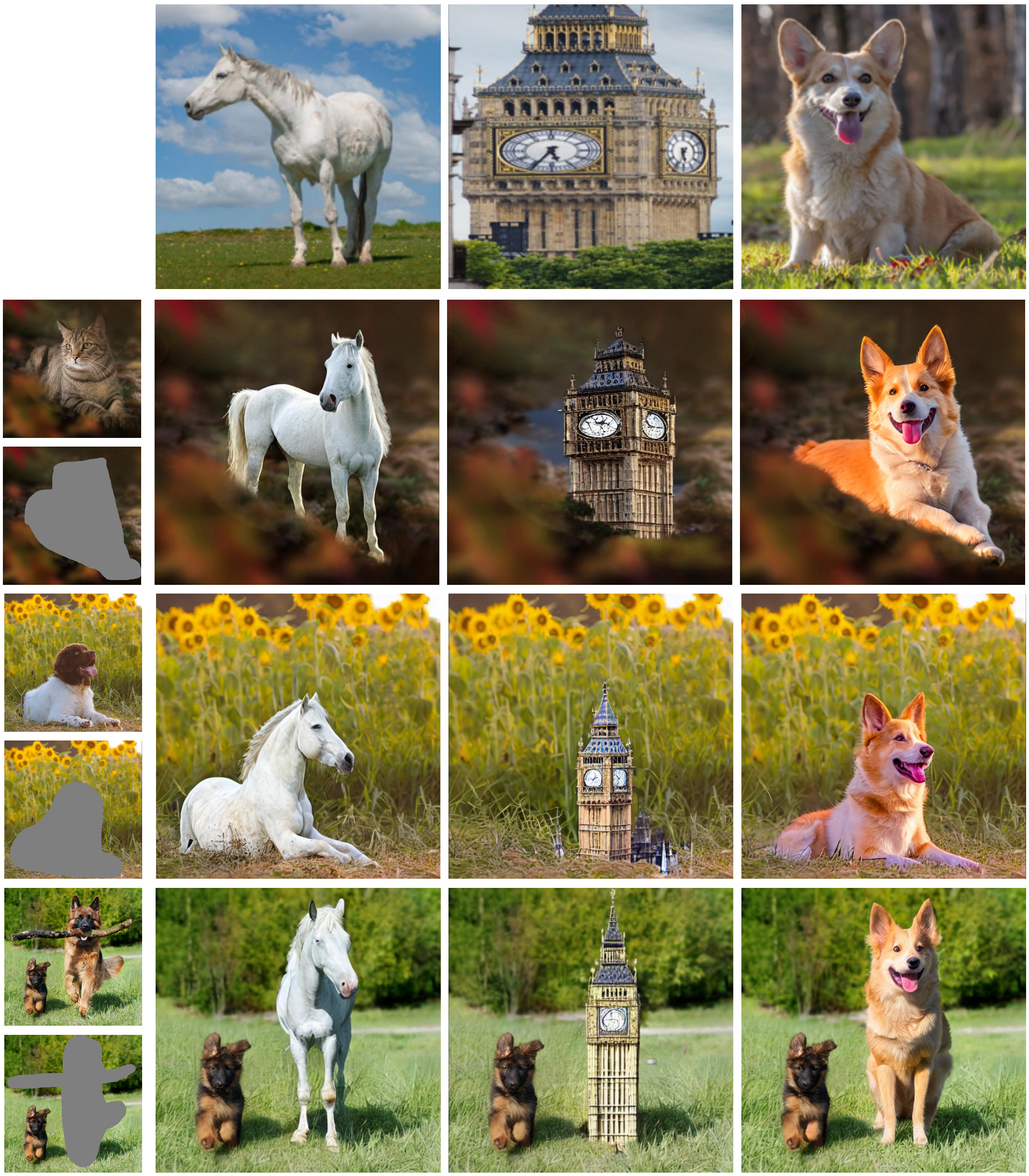}
\caption{Results of the same object with different source images. Our method is robust for different objects or different source images, even for some complicated objects, like 'Big Ben'.}
\label{fig:supp_1}
\end{figure*}

\section{Implementation details}
We adopt the Stable Diffusion~\cite{rombach2022high} as our baseline model and choose their publicly released v1-4 model for text-driven image generation as initialization. First, We modify it to a text-driven image inpainting model by expanding $5$ additional channels of the first convolution layer in the U-net (4 represents the encoded masked-image and 1 for the mask region). The new added weights are zero-initialized.
We choose CLIP~\cite{radford2021learning} pretrained model (ViT-L) as our image encoder and choose its feature from the last hidden state as condition. We utilize $15$ fully-connected (FC) layers to decode the feature from pretrained encoder and inject it into the diffusion process through cross attention.
We train the model using exponential moving average of weights and AdamW~\cite{loshchilov2017decoupled} optimizer with a constant learning rate of $1e-5$.
We use the HorizontalFlip ($p=0.5$), Rotate ($limit=20$), Blur ($p=0.3$) and ElasticTransform ($p=0.3$) from  Albumentations~\cite{info11020125} for image augmentation.
To quantitatively compare different settings, we adopt the CLIP image encoder (ViT-B) as feature extractor for FID, QS and CLIP Score. 
For the comparison with Stable Diffusion~\cite{rombach2022high}, we utilize their officially released code and pretrained text-driven inpainting model for testing (v1-5-inpainting).
For Blended Diffusion (image), we utilize CLIP (ViT-B) image encoder for encoding the reference images, which corresponds to its text encoder.

\section{Limitation}
Because the majority of the training data is natural photos, our method does not perform well with some artificial images, such as oil paintings. Furthermore, for some rarer objects like dinosaur, our method can hardly understand them well. We present some failure cases in Fig.~\ref{fig:supp_5}.

\begin{figure*}[h]
\centering
\includegraphics[width=0.8\columnwidth]{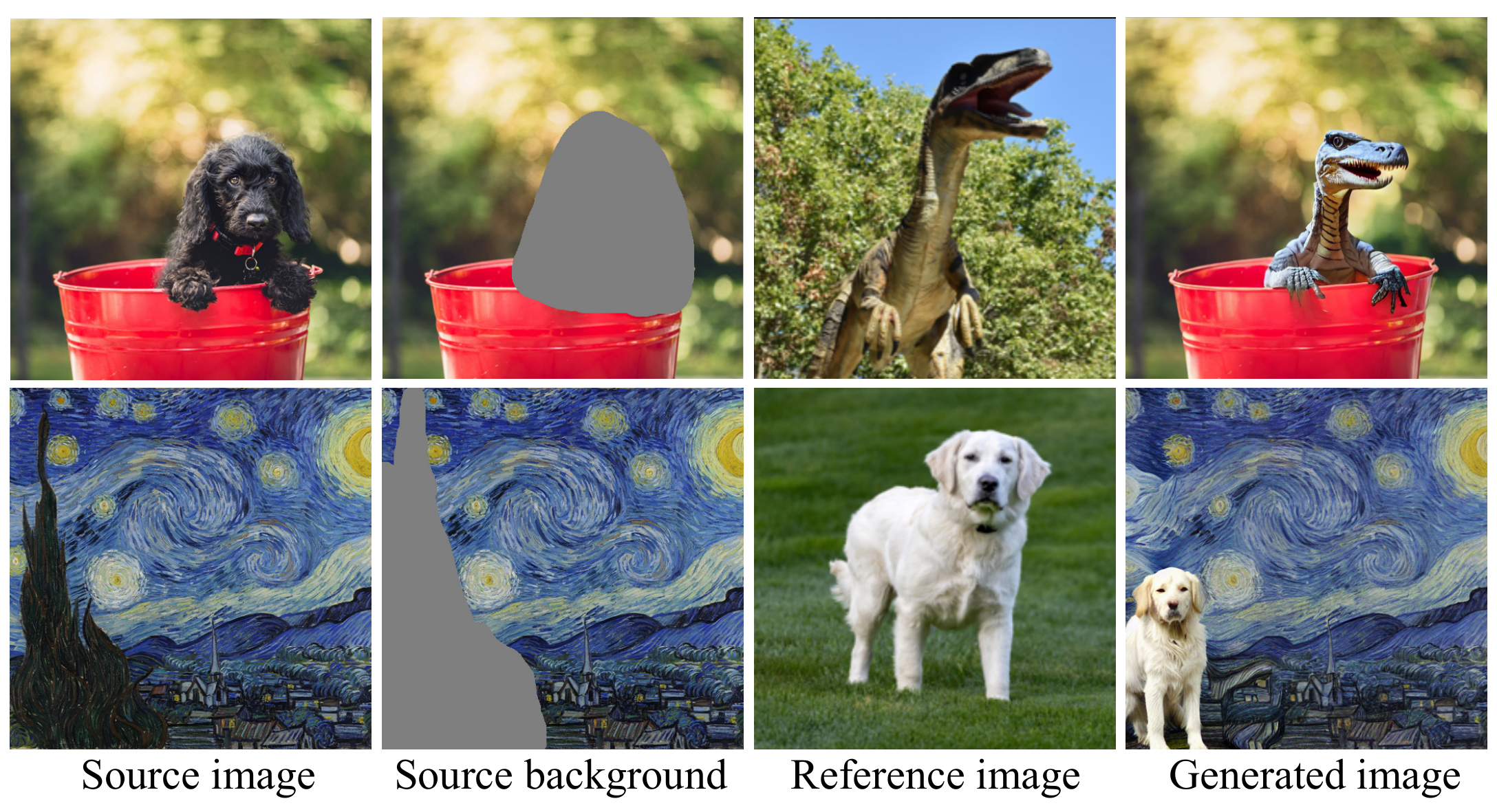}
\caption{Some failure cases.}
\label{fig:supp_5}
\end{figure*}
\clearpage
\end{appendices}
}

\end{document}